\documentclass[conference]{IEEEtran}
\IEEEoverridecommandlockouts
\usepackage{amsmath,amssymb,amsfonts}
\usepackage{algorithmic}
\usepackage{graphicx}
\usepackage{subfigure}
\usepackage{textcomp}
\usepackage{enumerate}
\usepackage[utf8x]{inputenc}
\usepackage{float}
\usepackage[square,numbers]{natbib}

\begin{document}
\title{Evaluation of trackers for Pan-Tilt-Zoom Scenarios}

\author{\IEEEauthorblockN{Yucao Tang\thanks{ This work was conducted while Yucao Tang was doing a MITACS Globalink internship at Polytechnique Montr\'{e}al}}
\IEEEauthorblockA{\textit{University of Electronic Science and Technology of China} \\
Chengdu, China \\
tang\_yucao@qq.com}
\and
\IEEEauthorblockN{Guillaume-Alexandre Bilodeau}
\IEEEauthorblockA{LITIV Lab. \\
\textit{Polytechnique Montr\'{e}al} \\
Montr\'{e}al, Canada \\
gabilodeau@polymtl.ca}
}

\maketitle

\begin{abstract}
Tracking with a Pan-Tilt-Zoom (PTZ) camera has been a research topic in computer vision for many years. Compared to tracking with a still camera, the images captured with a PTZ camera are highly dynamic in nature because the camera can perform large motion resulting in quickly changing capture conditions. Furthermore, tracking with a PTZ camera involves camera control to position the camera on the target. For successful tracking and camera control, the tracker must be fast enough, or has to be able to predict accurately the next position of the target. Therefore, standard benchmarks do not allow to assess properly the quality of a tracker for the PTZ scenario. In this work, we use a virtual PTZ framework to evaluate different tracking algorithms and compare their performances. We also extend the framework to add target position prediction for the next frame, accounting for camera motion and processing delays. By doing this, we can assess if predicting can make long-term tracking more robust as it may help slower algorithms for keeping the target in the field of view of the camera. Results confirm that both speed and robustness are required for tracking under the PTZ scenario. 
\end{abstract}

\section{Introduction}
Tracking with a Pan-Tilt-Zoom (PTZ) camera has been a research topic in computer vision for many years. Compared to tracking with a still camera, the images captured with a PTZ camera are highly changing in nature because the camera can perform large motion resulting in quickly changing capture conditions. Furthermore, tracking with a PTZ camera involves controlling a camera and it is an online process. Therefore, standard benchmarks, e.g. \cite{VOT}, do not allow to evaluate the performance of a tracker for the PTZ scenario because they do not account for camera control, neither for drop frames caused by spending too much time for processing a frame to track a target. A specific benchmark is required for evaluation in this scenario. In this work, we used the virtual PTZ framework that has been developed by \cite{chen} to evaluate recent trackers for the PTZ tracking scenario. The goal of this evaluation is to assess the performance of trackers in an online tracking scenario where a combination of speed and robustness is required. PTZ scenarios include many changes in scale and in illumination and large motion. We evaluated 19 tracking algorithms and compared and analyzed their performances. Many of these trackers were tested in the VOT 2016 \cite{VOT} benchmark. We also extended the framework to add target position prediction for the next frame, accounting for camera motion and processing delays. By doing this we can assess if prediction can help to make long-term tracking more robust and help slower algorithm for keeping the target in the field of view of the camera. 

\section{Evaluation framework and metrics}
Chen et al. \cite{chen} proposed a C++ framework to evaluate in a reproducible way trackers in PTZ scenarios. Because of the online nature of this scenario, the authors proposed the use of a PTZ camera simulator that pans, tilts and zooms based on a spherical video captured offline. The simulator also includes relevant delays that result in drop frames if tracking takes too long and if the target has a large motion in the image plane. These delays are categorized into execution delay, motion delay and communication delay.

\subsection{Simulator Configuration}
We use the evaluation framework as indicated by Chen et al. \cite{chen}. However, since some tracker codes are in C++ and others in Matlab, we adjusted how the delays are calculated to ensure fairness in execution time evaluation. We used the chronometer functions based on execution time in C++ 11 to calculate the time elapsed for processing frames by the trackers instead of the real-time clock. Some of the program running time is spent to read the image from disk drive and it was not included in the execution time. The motion delay is calculated by the time it takes for the simulated camera to tilt and/or pan. We decided that there would be no communication delays by supposing that the camera is not networked.

Some codes are written originally in Matlab and we had to use the Matlab engine to call the Matlab function, or in other word, tracker interfaces. Minor changes have been made to the Matlab source codes such as eliminating the display and drawing functions since those will affect the speed of processing but are totally unnecessary and are thus taken as irrelevant delays. Besides, after practical experiments, it turns out that calling Matlab engine from C++ is actually in a time scale of milliseconds and this overhead can be neglected since the time to process frame by trackers is much longer than that delay. 

\subsection{Performance Evaluation}\label{ssec:metrics}
Chen et al. defined four performance metrics to evaluate the trackers \cite{chen}.  These metrics are calculated in the image plane for the current camera viewpoint (i.e. the viewed subregion on the image sphere projected on the camera image plane). Let $C_{GT}$ and $C_{PT}$ be the ground-truth target center and the predicted target center, and $A_{GT}$ and $A_{PT}$ be the ground-truth target bounding box and the predicted bounding box, respectively. $C_{FOV}$ is the center of the camera image plane, or in other words, field of view (FOV). $TPE^t$ (Target Point Error) and $BOR^t$ (Box Overlapping Ratio) evaluate the quality of target localization and are defined as 
\begin{equation} 
TPE \text{ = } |C_{GT}-C_{PT}|
\end{equation}
and\\
\begin{equation} 
BOR^t \text{ = } \frac{A_{GT}\cap A_{PT}}{A_{GT}\cup A_{PT}}
\end{equation}  

$TPO^t$ (Target Point Offset) and $TF^t$ (Track Fragmentation) evaluate the quality of the camera control and are defined as
\begin{equation}  
TPO^t \text{ = } |C_{FOV}-C_{GT}|
\end{equation} 
and\\
\begin{equation}  
\text{$TF^t$ =}
\left\{  
             \begin{array}{lr}  
              1 , & \text{if $TPE^t$ is invalid}\\  
              0 , & \text{otherwise    }
             \end{array}
\right.  
\end{equation}  
$TF^t$ indicates whether the target is inside the camera FOV. $TPE^t$ and $TPO^t$ are invalid and assigned -1 if the target is outside the FOV. The overall metrics $TPE$, $TPO$, $BOR$ are the average metrics of the valid tracked frames. $TF$ is the sum of $TF^t$ divided by the number of processed frames. In the experiments, we report only $TF$ and $BOR$ as they are the most significant metrics. Besides $TPE$ and $BOR$ have similar purpose, and so are $TPO$ and $TF$.  

\section{Target Position Prediction}\label{sec:prediction}
In \cite{chen}, the authors made the remark that for a tracker to be successful, it should be either very fast, or should use some kind of target position prediction to keep the target close to the FOV of the camera. To assess the practicality of predicting a target position based on previous track information, we implemented three target motion models. Since there will be a delay between frames, it is necessary to predict the object position on the image sphere and move the camera accordingly so that the target appears in its FOV. Therefore, the prediction must account for the motion of the target and the motion of the camera. For calculation, let a target in the first frame appear at point $P_0$ (on the spherical image). At next frame, the target moves to $P_1$ (again on the spherical image). We can calculate its speed as $V_0=P_1-P_0/t_1-t_0$. Then in the third frame if the target moved from $P_1$ to $P_2$ its speed would then be $V_1=P_2-P_1/t_2-t_1$. Thus a basic classical mechanics model can be used to estimate the next position in the fourth frame.

The position prediction should locate the target near the image center as much as possible. By knowing the motion of the target it is possible to predict where it will be later after a delay $\Delta_t$.  $\Delta_t$ should account for the processing time of the current frame in addition to the time it takes for the camera to move. We experimented with three motion models to obtain the target motion ($\Delta_d$) between two instants:

\begin{itemize}
\item Model 1: Object is moving at a constant speed and uses the velocity of last instant
\begin{equation}
\Delta_d=V_1\times \Delta_t
\end{equation} 
\item Model 2: Object is moving at a constant speed and uses mean velocity in last two instant.
\begin{equation}
\Delta_d=\frac{(V_1+V_0)}{2} \times \Delta_t
\end{equation}
\item Model 3: Object can accelerate
\begin{equation}
\Delta_d=\frac{A\times {\Delta_t}^2}{2}+ V_1 \times \Delta_t
\end{equation}
where $A=V_1-V_0/t_1-t_0$ is the acceleration.
\end{itemize}

\section{Tested Trackers}
Among the 19 trackers we tested, six trackers are variations of correlation filters: KCF, SRDCF, SWCF, DSST, DFST and sKCF. Two trackers combine correlation filter outputs with color: STAPLE and STAPLE+. One is based on structured SVM: STRUCK. Two trackers are based purely on color: DAT and ASMS. One tracker is based on normalized cross-correlation: NCC. Two trackers are based on boosting: MIL and BOOSTING. One tracker is based on optical flow (MEDIANFLOW) and one tracker includes a detector (TLD). Two trackers can be categorized as part-based: DPCF and CTSE. Another one combines many trackers in an ensemble: KF-EBT. Below, we briefly describe the trackers. More details can be found in the original papers describing each of them. 

\begin{enumerate}
\item \textbf{Kernelized Correlation Filter tracker (KCF)} \cite{KCF} KCF is operating on HOG features. It localizes target with the equivalent of a kernel ridge regression trained with sample patches around the object at different translations. This version of the KCF tracker includes multi-scale support, sub-cell peak estimation and a model update by linear interpolation.
\item \textbf{Spatially Regularized Discriminative Correlation Filter Tracker (SRDCF)} \cite{SRDCF} This tracker is derived from KCF. It introduces a spatial regularization function that penalizes filter coefficients residing outside the target area, thus solving the problems arising from assumptions of periodicity in learning the correlation filters. The size of the training and detection samples can be increased without affecting the effective filter size. By selecting the spatial regularization function to have a sparse discrete Fourier spectrum, the filter is optimized directly in the Fourier domain. SRDCF employs also Color Names and greyscale features.
\item \textbf{Spatial Windowing for Correlation Filter-Based Visual Tracking (SWCF)} \cite{SWCF}
This tracker is derived from KCF. It predicts a spatial window for the observation of the object so that the correlation output of the correlation filter as well as the windowed observation are improved. Moreover, the estimated spatial window of the object patch indicates the object regions that are useful for correlation. 
\item \textbf{Discriminative Scale Space Tracker (DSST)} \cite{DSST} DSST extends the Minimum Output Sum of Squared Errors (MOSSE)\cite{mosse} tracker with robust scale estimation. DSST also learns a one-dimensional discriminative scale filter which is used to predict the target size. The intensity features used in MOSSE \cite{mosse} tracker are combined with a pixel-dense representation of HOG features.
\item \textbf{Dynamic Feature Selection Tracker (DFST) \cite{DFST}}
DFST is a visual tracking algorithm based on the real-time selection of locally and temporally discriminative features. DFST provides a significant gain in accuracy and precision with respect to KCF by the use of a dynamic set of features.  A further improvement is given by making micro-shifts at the predicted position according the best template matching.
\item \textbf{Scalable Kernel Correlation Filter with Sparse Feature Integration (sKCF)} \cite{sKCF} This tracker is derived from KCF. It introduces an adjustable Gaussian window function and a keypoint-based model for scale estimation. It deals with the fixed window size limitation in KCF.
\item \textbf{Sum of Template And Pixel-wise LEarners (STAPLE)} \cite{Staple} STAPLE combines two image patch representations that are sensitive to complementary factors to learn a model that is robust to both color changes and deformations. It combines the scores of two models in a dense window translation search. The scores of the two models are indicative of their reliability.
\item \textbf{An improved STAPLE tracker with multiple feature integration (STAPLE+)} STAPLE+ is based on the STAPLE tracker and improves it by integrating multiple features. It extracts HOG features from color probability map to exploit color information better. The final response map is thus a fusion of scores obtained with different features.
\item \textbf{STRUCtured output tracking with Kernels (STRUCK) \cite{STRUCK}} This is a framework for adaptive visual object tracking. It applies a support vector machine which is learned online. It introduces a budgeting mechanism that prevents the unbounded growth in the number of support vectors that would otherwise occur during tracking.
\item \textbf{Distractor Aware Tracker (DAT)} \cite{DAT} This is a tracking-by-detection approach based on appearance. To distinguish the object from the surrounding areas, a discriminative model using color histograms is applied. It adapts the object representation beforehand so that distractors are suppressed and the risk of drifting is reduced.
\item \textbf{Scale Adaptive Mean Shift (ASMS)} \cite{ASMS} This is a mean-shift tracker \cite{meanshift} optimizing the Hellinger distance between a template histogram and the target in the image. The optimization is done by a gradient descent. ASMS addresses the problem of scale adaptation and scale estimation. It also introduces two improvements over the original mean-shift \cite{meanshift} to make the scale estimation more robust in the presence of background clutter: 1) a histogram color weighting and 2) a forward-backward consistency check.
\item \textbf{Normalized Cross-Correlation (NCC)} \cite{ncc} NCC follows the basic idea of tracking by searching for the best match between a static grayscale template and the image using normalized cross-correlation.
\item \textbf{Multiple Instance Learning tracker (MIL) }\cite{MIL} MIL uses a tracking-by-detection approach with multiple instance learning instead of traditional supervised learning methods. It shows improved robustness to inaccuracies of the tracker and to incorrectly labeled training samples. 
\item \textbf{BOOSTING}  \cite{boosting} It is based on MIL\cite{MIL}. This is a real-time object tracker based on a novel on-line version of the AdaBoost algorithm. The classifier uses the surrounding background as negative examples in the update step to avoid the drifting problem. 
\item \textbf{MEDIANFLOW} \cite{MEDIANFLOW} This tracker uses optical flow to match points between frames. The tracking is performed forward and backward in time and the discrepancies between these two trajectories are measured. The proposed error enables reliable detection of tracking failures and selection of reliable trajectories in video sequences.
\item \textbf{Tracking-learning-detection(TLD) }\cite{TLD} It combines both a tracker and a detector. The tracker follows the object from frame to frame using MEDIANFLOW \cite{MEDIANFLOW}. The detector localizes the target using all appearances that have been observed so far and corrects the tracker if necessary. The learning estimates the detector errors and updates it to avoid these errors in the future.
\item \textbf{Deformable Part-based Tracking by Coupled Global and Local Correlation Filters (DPCF)} \cite{DPCF}
This tracker that is derived from KCF relies on joint interactions between a global filter and local part filters. The local filters provide an initial estimate which is used by the global filter as a reference to determine the final result. The global filter provides a feedback to the part filters. In this way, it handles both partial occlusion (with the part filters) but also scale changes (with the global filter).
\item \textbf{Contextual Object Tracker with Structural Encoding (CTSE)}\cite{CTSE} This tracker uses contextual and structural information (that is specific to a target object) into the appearance model. This is first achieved by including features from complementary region having correlated motion with the target object. Second, a local structure that represents the spatial constraints between features within the target object is included. SIFT keypoints are used as features to encode both these information. 
\item \textbf{Kalman filter ensemble-based tracker (KF-EBT)} This tracker combines the result of two other trackers: ASMS\cite{ASMS} using a color histogram and KCF. Using a Kalman filter, the tracker works in cycles of prediction and correction. Firstly, a motion model predicts the target next position. Secondly, the trackers results are fused with the predicted position and the model is updated.
\end{enumerate}

\section{Results and analysis}
In this section, we report our results on the 36 video sequences of \cite{chen}. The video sequences that consist in tracking persons, faces and objects include difficulties such as motion blur, scale change, out-of-plane rotation, fast motion, cluttered background, illumination variation, low resolution, occlusion, presence of distractors and articulated objects. Results are reported for the whole dataset.

\subsection{Ranking method}
During testing, we discovered that since different trackers have different tracking speed, using only the four metrics in section \ref{ssec:metrics} is not enough. For example, some trackers have $TPE$, $TPO$ and $BOR$ metrics the are good because they track slowly in real-time simulation, which means they just track a few frames correctly and all other frames are invalid and are ignored for the calculation of the metrics. Under this circumstance, the tracker succeed in tracking every processed frames (which are only the first frames). Only $TF$ can capture to some extent this lack of performance as it verifies if the target is in the FOV or not. Thus we consider another essential metric: processed frame ratio ($PR$):
\begin{equation} 
PR=\frac{F_{NP}}{F_{TO}},
\end{equation} 
where $F_{NP}$ is the number of processed frames and $F_{TO}$ is the total number of frames.

$TF$ contains part of $PR$ information since it stands for whether the object is inside the FOV in the processed frames. If $PR$ is low, $TF$ will be high since the tracker will not be able to track the object correctly in the processed frames because of the long interval between them. However, a high $TF$ can be caused also by poor robustness. 

After considering the PTZ camera nature and the tracker test results, we formulated a ranking formula. The formula stands for the Euclidean distance between the point defined by the pair ($BOR$, $TF$) obtained by a tracker and the ideal tracker (top-right point in figure \ref{figure1}). The score is thus:
\begin{equation} 
Score=\sqrt[]{(1-BOR)^2+TF^2}.
\label{eq:score}
\end{equation} 
We selected $TF$ and $BOR$ because we consider that $TPE$ conveys similar information as $BOR$ and $TF$ conveys similar information as $TPO$.

\subsection{Results without processing delays}
We have first set the execution ratio to zero in order to compare different trackers for their performances for in-plane rotations, out-of-plane rotations and drastic scene changes caused by the camera motion. We are looking for the most robust trackers, neglecting their processing times (they are set to all perform at the same speed). Note that in this experiment, the camera motion delays are considered as they reflects the robustness of the trackers. If a tracker performs poorly, this will cause unnecessary camera motion that will result in drop frames.  

Table \ref{table:exec0} and Figure \ref{figure:exec0} give the results for the 19 trackers based on their ranking. In the PTZ camera scenario, the difficulties with in-plane and out-of-plane rotations will be amplified because of the application dynamic nature. Surprisingly, trackers which adopted a scale adaptation function such as SRDCF do not necessarily perform better than other trackers due in great part to their slow execution speed. ASMS and DPCF are the best performers in this experiment. In the VOT 2016 benchmark \cite{VOT} there is no drop frames caused by delays and less viewpoint changes caused by camera motion. Thus it is reasonable that there is difference between the ranking of our framework and that of the VOT ranking. However, our ranking is still quite similar to that of VOT 2016 benchmark. Trackers like Staple, Staple+, KCF-EBT and DAT are good both in VOT benchmark and our benchmark. However, in our benchmark the performance of ASMS is surprisingly the best when it just ranked in the middle in VOT. And some trackers like SRDCF do not behave well in our PTZ framework probably because they do not output bounding boxes when the tracking fails and as a result, the PTZ camera is not controlled correctly. The VOT system will check every five frames to verify whether the tracker has failed. If it has failed, it is reinitialized. In our framework, we do not intervene in the tracking process at all. Early failures are thus more penalized.
\begin{table}
\centering
\caption{Tracker performances with execution ratio set to 0 on the 36 video sequences of \cite{chen}. Trackers are ranked based on their score (equation \ref{eq:score}).}
\label{table:exec0}
\begin{tabular}{lllll}
\hline
Tracker Names & Score & $BOR$ & $TF$ & $PR$ \\ \hline
ASMS & 0.65 & 0.42 & 0.30 & 0.92 \\ \hline
DPCF {[}MATLAB{]} & 0.71 & 0.40 & 0.38 & 0.80 \\ \hline
TLD & 0.72 & 0.56 & 0.57 & 0.41 \\ \hline
DAT {[}MATLAB{]} & 0.72 & 0.35 & 0.31 & 0.77 \\ \hline
Staple+ {[}MATLAB{]} & 0.72 & 0.37 & 0.35 & 1.00 \\ \hline
Staple {[}MATLAB{]} & 0.72 & 0.36 & 0.34 & 1.00 \\ \hline
KF-EBT & 0.74 & 0.36 & 0.37 & 0.98 \\ \hline
DSST & 0.78 & 0.38 & 0.48 & 0.92 \\ \hline
STRUCK & 0.85 & 0.31 & 0.50 & 0.94 \\ \hline
SKCF & 0.87 & 0.30 & 0.52 & 0.91 \\ \hline
KCF & 0.88 & 0.28 & 0.51 & 0.95 \\ \hline
SWCF {[}MATLAB{]} & 0.88 & 0.31 & 0.55 & 0.75 \\ \hline
MIL & 0.89 & 0.28 & 0.52 & 0.85 \\ \hline
DFST {[}MATLAB{]} & 0.90 & 0.27 & 0.53 & 0.93 \\ \hline
BOOSTING & 0.92 & 0.26 & 0.54 & 0.62 \\ \hline
MEDIANFLOW & 0.98 & 0.27 & 0.65 & 0.95 \\ \hline
SRDCF {[}MATLAB{]} & 1.01 & 0.73 & 0.97 & 0.14 \\ \hline
NCC & 1.03 & 0.24 & 0.69 & 0.90 \\ \hline
CTSE & 1.21 & 0.04 & 0.74 & 0.44 \\ \hline
\end{tabular}
\end{table}

\subsection{Results with processing delays}
\begin{figure*}
\centering
\subfigure[Without processing delay]
{
  \includegraphics[trim={7cm 0 7cm 0},clip,width=0.45\textwidth]{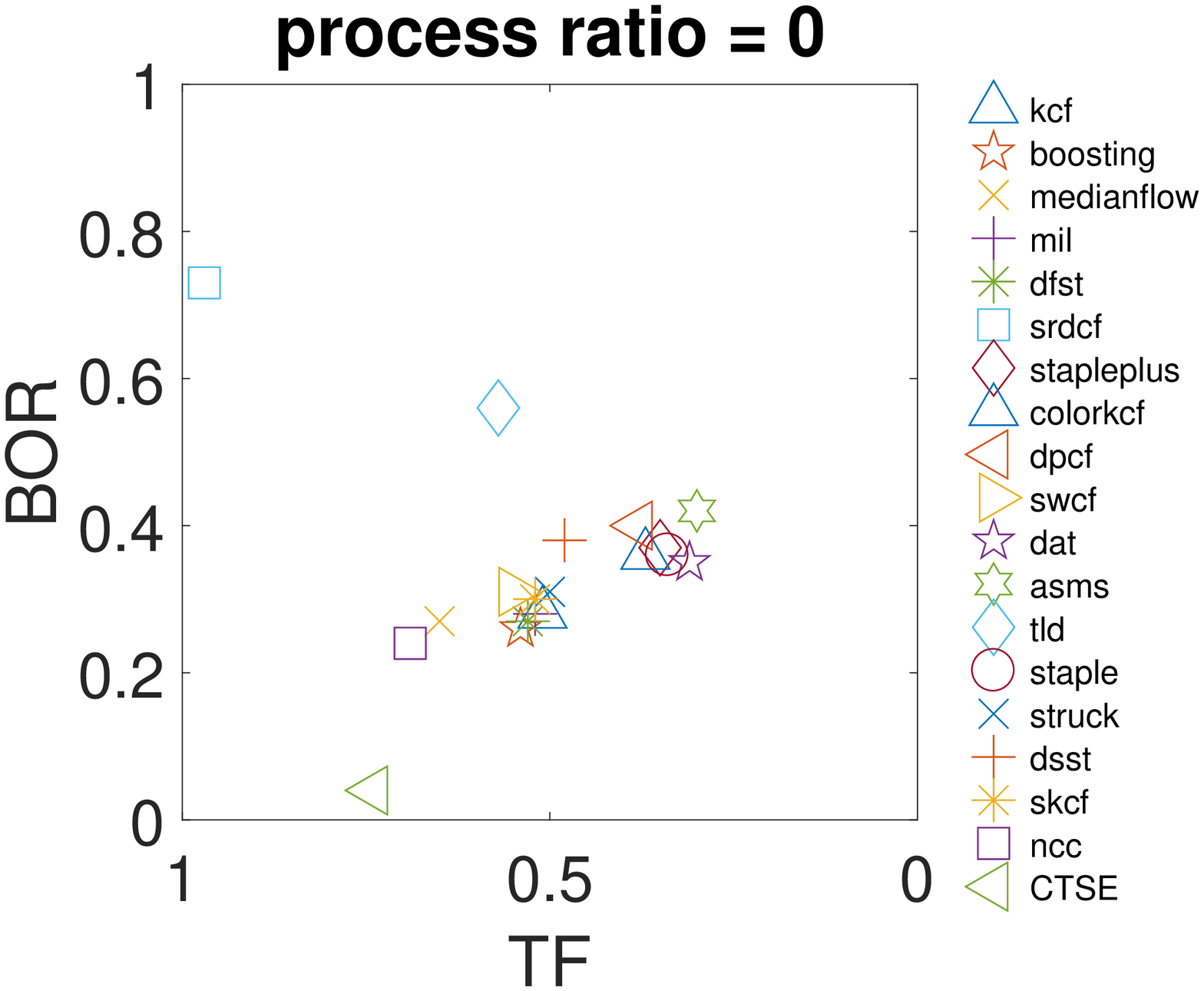}
  \label{figure:exec0}
}
\subfigure[With processing delay]
{
  \includegraphics[trim={7cm 0 7cm 0},clip,width=0.45\textwidth]{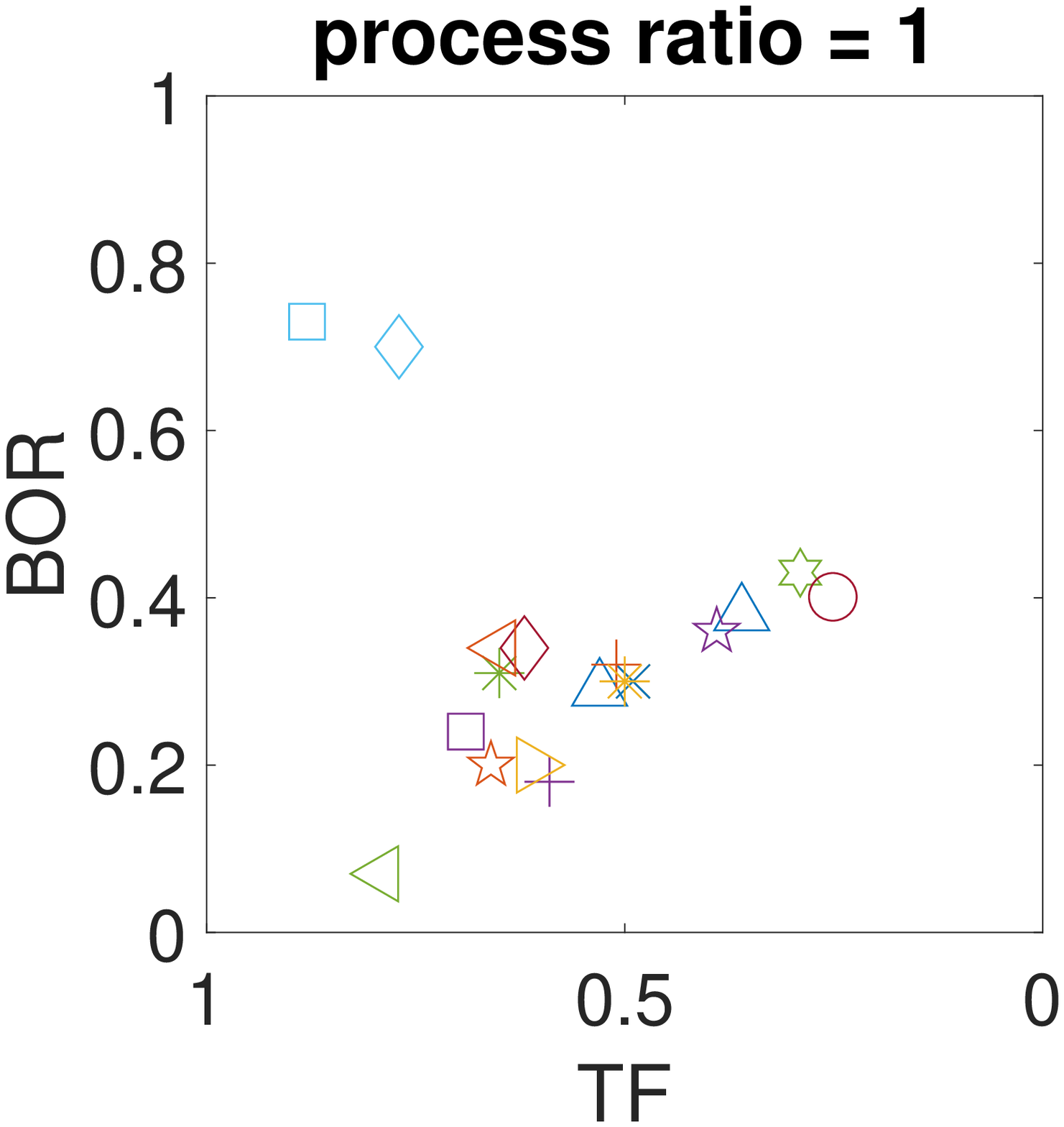}
  \label{figure:exec1}
}
\caption{$BOR$ vs $TF$ scatter plot without (left) and with processing delays (right) on the 36 video sequences of \cite{chen}. The legend that indicates the symbols representing the trackers is the same for both plots. Ideal results are $BOR=1$ and $TF=0$.}
\label{figure1}
\end{figure*}

We then set the execution ratio to 1 to track objects. In the real world the objects will keep moving while the tracker is processing a frame. As a result, the task of tracking is harder in this case since the intervals between processed frames are caused both by the time for processing the frame and the camera motion delay. $PR$ should decline and the trackers should lose targets more easily. Table \ref{table:exec1} and Figure \ref{figure:exec1} give the results for the 19 trackers.

Compared to the previous case, the ranking of trackers changes. ASMS still ranks first, but DPCF degrades because it is very slow and its $PR$ value declines to 0.08 where it was previously at 0.8. The other trackers relative rankings do not change much, but the average score of trackers is higher, which means that their performances are worse because of the execution delay in the tracking process. Still, compared to the VOT 2016 benchmark \cite{VOT}, the performance of ASMS is still surprisingly the best. This means that this tracker is particularly good for handling viewpoint changes. Finally, performance in VOT are better than for our task because we are testing online tracking and camera control. The PTZ scenario requires tracking people with different illumination variation, different scale and from rapidly changing viewpoints. All of those reasons make our results unique compared to other benchmarks.

\begin{table}
\centering
\caption{Tracker performances with execution ratio set to 1 on the 36 video sequences of \cite{chen}. Trackers are ranked based on their score (equation \ref{eq:score}).}
\label{table:exec1}
\begin{tabular}{lllll}
\hline
Tracker Names & Score & $BOR$ & $TF$ & $PR$ \\ \hline
ASMS & 0.64 & 0.43 & 0.29 & 0.88 \\ \hline
Staple {[}MATLAB{]} & 0.65 & 0.40 & 0.25 & 0.35 \\ \hline
KF-EBT & 0.72 & 0.38 & 0.36 & 0.71 \\ \hline
DAT {[}MATLAB{]} & 0.75 & 0.36 & 0.39 & 0.21 \\ \hline
TLD & 0.83 & 0.70 & 0.77 & 0.13 \\ \hline
DSST & 0.85 & 0.32 & 0.51 & 0.48 \\ \hline
KCF & 0.85 & 0.30 & 0.49 & 0.63 \\ \hline
STRUCK & 0.85 & 0.30 & 0.49 & 0.15 \\ \hline
SKCF & 0.86 & 0.30 & 0.50 & 0.90 \\ \hline
BOOSTING & 0.89 & 0.29 & 0.53 & 0.54 \\ \hline
Staple+ {[}MATLAB{]} & 0.91 & 0.34 & 0.62 & 0.08 \\ \hline
SRDCF {[}MATLAB{]} & 0.92 & 0.73 & 0.88 & 0.03 \\ \hline
DPCF {[}MATLAB{]} & 0.93 & 0.34 & 0.65 & 0.08 \\ \hline
DFST {[}MATLAB{]} & 0.95 & 0.31 & 0.65 & 0.09 \\ \hline
MEDIANFLOW & 0.96 & 0.27 & 0.62 & 0.85 \\ \hline
SWCF {[}MATLAB{]} & 1.01 & 0.20 & 0.61 & 0.23 \\ \hline
NCC & 1.03 & 0.24 & 0.69 & 0.88 \\ \hline
MIL & 1.04 & 0.20 & 0.66 & 0.15 \\ \hline
CTSE & 1.22 & 0.07 & 0.79 & 0.11 \\ \hline
\end{tabular}
\end{table}

\subsection{Target Position Prediction}
Finally, we tested target position prediction to investigate if it can help trackers to perform better. Table \ref{prediction} gives results for two trackers. Results are similar for all the other trackers. The proposed models (see section \ref{sec:prediction}) for predicting the next position of the target are not improving results. This is due to the fact that the speed of the target is difficult to estimate because it moves in 3D, but we estimate its motion in 2D. Therefore, the predicted speed is not very accurate. After calculating the speed, the framework will use this speed to predict the target position in the next frame. It may cause unnecessary large motion by the camera. For example, the predicted target motion may be too large so the camera, by rotating to this wrongly predicted position, will add delays in the tracking process. This adds to the possibility that the tracker will lose the target. If the target cannot be tracked, its speed will not be updated leading to an even worse situation where the camera just rotate more or less randomly. The high-speed trackers are affected the most by wrong prediction. Their process ratio decline from above 0.8 to below 0.2. 

Therefore, we can conclude from this experiment that although appealing in theory, compensating slow tracking by a position prediction is not easy to apply in practice. It may work for objects that are far away and that mostly move in a plane, but it cannot work for target that are closer and that are moving toward or away from the camera. In such cases, the motion of the target cannot be predicted in 2D. For best results, it is thus preferable to design a fast tracker. 

\begin{table}
\centering
\caption{Tracker performances with execution ratio set to 1 and with various prediction methods on the 36 video sequences of \cite{chen}.}
\label{prediction}
\begin{tabular}{llllll}
\hline
Tracker Names & Prediction & Score & $BOR$ & $TF$ & $PR$ \\ \hline
KCF &  None & 0.85 & 0.30 & 0.49 & 0.63 \\ \hline
KCF & Model 1& 1.06  & 0.20 & 0.69 & 0.14 \\ \hline
KCF & Model 2& 0.97  & 0.24 & 0.61 & 0.15 \\ \hline
KCF & Model 3 & 1.05  & 0.24 & 0.73 & 0.12 \\ \hline
BOOSTING & None & 0.89  & 0.29 & 0.53 & 0.54 \\ \hline
BOOSTING & Model 1 & 1.03  & 0.26 & 0.71 & 0.12 \\ \hline
BOOSTING & Model 2 & 1.10  & 0.17 & 0.72 & 0.12 \\ \hline
BOOSTING & Model 3 & 1.08  & 0.18 & 0.70 & 0.11 \\ \hline

\end{tabular}
\end{table}

\section{Conclusion}
This paper presented a benchmark of recent trackers for the PTZ tracking scenario. Surprisingly, high-speed trackers, such as MEDIANFLOW and NCC, do not necessarily behave better than others. However, since predicting the target position was shown to be difficult, slow trackers should be avoided for the PTZ tracking task.  
The results of our test indicate that the top performing tracker for the PTZ scenario is the ASMS tracker. This tracker performed very well in accuracy as well as in robustness in our tests. It is impossible to conclusively determine whether the performance of ASMS over other trackers come from its image features or its approach. Nevertheless, results of top trackers show that features play a significant role in the final performance.
\bibliographystyle{IEEEtran} % or try abbrvnat or unsrtnat
\bibliography{tycbib} % refers to example.bib

\end{document}